\documentclass{article} 
\usepackage{iclr2020_conference,times}

\usepackage{amsmath,amsfonts,bm}

\def\eqref#1{equation~\ref{#1}}

\def\1{\bm{1}}

\DeclareMathAlphabet{\mathsfit}{\encodingdefault}{\sfdefault}{m}{sl}
\SetMathAlphabet{\mathsfit}{bold}{\encodingdefault}{\sfdefault}{bx}{n}

\usepackage{hyperref}
\usepackage{url}

\title{Igbo-English Machine Translation:\\An Evaluation Benchmark}

\author{I. Ezeani, P. Rayson
\\
Lancaster University,\\
Lancaster, UK\\
\And
I. Onyenwe, C. Uchechukwu \\
Nnamdi Azikiwe University, \\
Awka, Nigeria \\
\And
Mark Hepple \\
Sheffield University, \\
Sheffield, UK \\
}
\iclrfinalcopy 
\begin{document}
\maketitle

\section{Introduction}
Although researchers are pushing the boundaries and enhancing the capacities of NLP tools and methods, works on African languages are lagging behind. A lot of focus on well-resourced languages such as English, Japanese, German, French, Russian, Mandarin Chinese etc. Over 97\% of the world’s 7000 languages, including African languages, are low-resourced for NLP i.e. they have little or no data, tools, and techniques for NLP research. For instance, only 5 out of 2965 (0.19\%) authors of full-text papers in the ACL Anthology\footnote{\url{https://www.aclweb.org/anthology/}} extracted from the 5 major conferences in 2018 (ACL, NAACL, EMNLP, COLING and CoNLL) are affiliated to African institutions\footnote{\textbf{Source:}  \url{http://www.marekrei.com/blog/geographic-diversity-of-nlp-conferences/}}.

In this work, we discuss our effort toward building a standard evaluation benchmark dataset for Igbo-English machine translation tasks. Igbo\footnote{\textbf{Igbo:} \url{https://en.wikipedia.org/wiki/Igbo_language}} is one of the 3 major Nigerian languages spoken by over 50 million people globally, 50\% of whom are in southeastern Nigeria. Igbo is low-resourced despite some efforts toward developing IgboNLP such as part-of-speech tagging: \cite{onyenwe2014part}, \cite{onyenwe2019toward}; and diacritic restoration: \cite{ezeani2016automatic}, \cite{ezeani2018transferred}.

Although there are exiting sources for collecting Igbo monolingual and parallel data, such as the \textit{OPUS Project} (\cite{TIEDEMANN12.463}) or the \url{JW.ORG}, they have certain limitations. The OPUS Project is a good source training data but, given that there are no human validations, may not be good as an evaluation benchmark. \url{JW.ORG} contents, on the other hand, are human generated and of good quality but the genre is often skewed to religious contexts and therefore may not be good for building a generalisable model.

This project focuses on creating and publicly releasing a standard evaluation benchmark dataset for Igbo-English machine translation research for the NLP research community. This project aims to build, maintain and publicly share a standard benchmark dataset for Igbo-English machine translation research . There are three key objectives:
\begin{enumerate}
    \item Create a minimum of 10,000 English-Igbo human-level quality sentence pairs mostly from the news domain
    \item To assemble and clean a minimum of 100,000 monolingual Igbo sentences, mostly from the news domain, as companion monolingual data for training MT models
    \item To release the dataset to the research community as well as present it at a conference and publish a journal paper that details the processes involved.
\end{enumerate}

\section{Methods}
To achieve the objectives above, the task was broken down in the following phases:\\
\\
\textbf{Phase 1: Raw data collection and pre-processing:}\\ This phase is to produce cleaned and pre-processed a minimum 10,000 sentences: 5,000 English and 5,000 Igbo. It involved the collection, cleaning and pre-processing (normalisation, diacritic  restoration, spelling correction etc.) of Igbo and English sentences from freely available electronic texts (e.g. Wikipedia, CommonCrawl, local government materials, local TV/Radio stations etc).\\
\\
\textbf{Phase 2: Translation and correction}\\
In this phase, the 10,000 sentence pairs are created manual translation and correction. The key tasks include:
\begin{enumerate}
    \item Translating English sentences to Igbo (EN-IG)
    \item Translating Igbo sentences to English (IG-EN)
    \item Correcting the translations
\end{enumerate}
5 Igbo speakers were engaged for the bidirectional of translations while 3 other Igbo speakers, including an Igbo linguist are assisting with the on-going corrections. Chunks ($\approx$ 250 each) of sentences are given to each translator in each direction (i.e. IG-EN and EN-IG). At the time of submission, we have $11,584$ sentence pairs as detailed in Table \ref{breakdown} while the splits of the parallel data into \textit{development}, \textit{text} and \textit{hidden test} sets is shown in Table \ref{splits}

\begin{table}[!ht]
\caption{Breakdown of the Benchmark Evaluation Parallel Data}
\label{breakdown}
\centering
\begin{tabular}{lrl}
\hline
\multicolumn{1}{l}{\textbf{Type}}&\multicolumn{1}{c}{\textbf{Sent pairs}}&\multicolumn{1}{l}{\textbf{Sources}}\\
\hline
\textit{Igbo-English} & 5,836 & \url{https://www.bbc.com/igbo}\\
\textit{English-Igbo} & 5,748 & Mostly from local newspapers (e.g. Punch)\\
\hline
\textit{Total} & $11,584$ &\\
\hline
\end{tabular}
\caption{Splits of the Benchmark Evaluation Parallel Data}
\label{splits}
\begin{tabular}{lrl}
\hline
\textbf{Evaluation Splits} & \textbf{IG-EN} & \textbf{EN-IG}\\
\hline
\textit{Development Set} & 5000 & \multicolumn{1}{r}{5000}\\
\textit{Test set} & 500 & \multicolumn{1}{r}{500}\\
\textit{Hidden Test} & 336 & \multicolumn{1}{r}{248}\\
\hline
\end{tabular}
\end{table}


\textbf{Phase 3: Manual checks and Inter-translator Agreement}\\
This phase is currently on-going and it involves manually checking and correcting the 10,000 translated sentence pairs. This is to ensure that the translations conform with the contemporary communicative usage of the languages. Our approach so far is simplistic i.e. it seeks to establish absolute agreement between translators. We know it could overstate agreement (\cite{lommel2014assessing}), but we believe it will improve the quality of the translation. More work will be done in this area in future.\\
\linebreak
\textbf{Phase 4: Monolingual Igbo sentence collection and pre-processing}\\
The aim here is to collect and clean a minimum of 100,000 monolingual Igbo sentences. the cleaning process involves normalisation, diacritic restoration, spelling correction from freely available sources (news, government materials, Igbo literature, local TV/Radio stations etc).

A large chunk of the data is collected from the Jehova's Witness Igbo\footnote{\textbf{Source:} \url{https://www.jw.org/ig/}} contents. Though we included the Bible, more contemporary contents (books and magazine e.g. \textit{Teta!} (\textit{Awake!}), \textit{Ulo Nche!} (\textit{WatchTower})) were the main focus. Also, we got contents from BBC-Igbo\footnote{\url{https://www.bbc.com/igbo/}} and Igbo-Radio (\url{https://www.bbc.com/igbo}) as well as Igbo literary works(\textit{Eze Goes To School}\footnote{\url{https:// bit.ly/2vdGvKN}} and \textit{Mmadu Ka A Na-Aria} by Chuma Okeke). This phase is still on-going but we have so far collected and cleaned $\approx$ 380k Igbo sentences as detailed in Table \ref{data_sources_counts}. It is important to point out that we have also collected data in other formats (e.g. audio, non-electronic texts) from local media houses which we hope to also transcribe and include in our collection.

\begin{table}[h!]
\caption{Data Sources and Counts}
\label{data_sources_counts}
\centering
\begin{tabular}{lrrr}
\multicolumn{1}{c}{\bf Source}  &\multicolumn{1}{c}{\bf Sentences} &\multicolumn{1}{c}{\bf Tokens} &\multicolumn{1}{c}{\bf UniqToks}\\
\hline
eze-goes-to-school.txt  &   1272 & 25413 & 2616\\
mmadu-ka-a-na-aria.txt  &   2023 & 39731 & 3292\\
bbc-igbo.txt            &  34056 & 566804 & 28459\\
igbo-radio.txt          &   5131 & 191450 & 13391\\
jw-ot-igbo.txt          &  32251 & 712349 & 13417\\
jw-nt-igbo.txt          &  10334 & 253806 & 6731\\
jw-books.txt            & 142753 & 1879755 & 25617\\
jw-teta.txt             &  14097 & 196818 & 7689\\
jw-ulo-nche.txt         &  27760 & 392412 & 10868\\
jw-ulo-nche-naamu.txt   & 113772 & 1465663 & 17870\\
\hline
\multicolumn{1}{c}{\bf Total} & \textbf{383,449} & \textbf{5,724,201} & \textbf{69,091}\\
\hline
\end{tabular}
\end{table}

\section{Access to data}
All data generated as described above are available under the Creative Commons license from this GitHub repository\footnote{\url{https://github.com/IgnatiusEzeani/IGBONLP/tree/master/ig_en_mt}} and will be regularly updated.

\section{Conclusion}
This work presents an on-going project on building a benchmark evaluation dataset for Igbo--English machine translation project. The released dataset will hopefully be useful in fairly and more reliably comparing the performance of models built for IG-EN translations.

Our efforts in increasing the size of the sentence pairs as well as improving the quality of translations will continue in will be published as we progress. In addition to releasing the dataset to the research community, our plan for future works include building and comparing various machine translation models based on the current state-of-the-art methods. This will be followed by an in-depth analysis of their performances. 

\subsubsection*{Acknowledgments}
The authors wish to acknowledge and thank Facebook AI Research (Facebook AI) for funding this project. Our immense gratitude also goes to Marc’Aurelio Ranzato and Francisco Guzm\'an for initiating, facilitating the funding and providing us with a lot of technical ideas.

\bibliography{ig_en_mt}
\bibliographystyle{iclr2020_conference}

\end{document}